\documentclass[conference]{IEEEtran}

\usepackage{cite}
\usepackage{amsmath,amssymb,amsfonts}
\usepackage{graphicx}
\usepackage{booktabs}
\usepackage{multirow}
\usepackage{textcomp}
\usepackage{hyperref}
\usepackage{xcolor}
\usepackage{subfloat}
\usepackage{subfig}

\usepackage{algorithm}
\usepackage{algpseudocode}
\usepackage{amsmath}
\usepackage{amssymb}

\newcommand{\moxie}{MOXIE}
\newcommand{\ime}{IME}

\begin{document}

\title{MOXIE: Discovering Alternative Explanations for Biomedical Image Classifiers}

\author{
\IEEEauthorblockN{Abiha Tahsin Chowdhury, Rahul Dubey}
\IEEEauthorblockA{Department of Computer Science, Missouri State University, Springfield, MO, USA\\
Email: RahulDubey@MissouriState.edu}
}

\maketitle

\begin{abstract}
Segment-based explanation methods such as LIME return a single explanation for each prediction, computed from one fixed image segmentation. This hides two important facts: a prediction can be supported by many different sets of image segments, and the segmentation itself shapes which explanations can be found. We introduce MOXIE (Multi-Objective eXplanation Imaging Engine), an evolutionary framework that searches for segment subsets that preserve the classifier's confidence while keeping as little of the image as possible. Instead of one explanation, MOXIE returns a Pareto front of alternative explanations that range from compact to highly faithful. We evaluate MOXIE with NSGA-II and four segmentation methods (SLIC, Felzenszwalb, Watershed and Voronoi) on BloodMNIST and HAM10000 datasets, using the same evaluation budget as LIME. Results show that MOXIE achieves a higher hypervolume than LIME on every image. LIME's explanations often appear convincing, yet the classifier's confidence collapses when only the highlighted segments are shown. MOXIE’s fronts reveal how much of the image is needed to preserve the model’s confidence and which contextual regions influence it. We also find that segmentation strongly affects evaluation: methods with unequal segment sizes appear most compact when segments are counted. These results show that alternative explanations provide a more complete view of a model’s decision than a single explanation.
\end{abstract}


\section{Introduction}
\label{sec:intro}

Deep learning has demonstrated strong performance in medical image analysis, including Alzheimer's disease (AD) assessment from magnetic resonance imaging (MRI), blood-cell classification, and radiographic screening~\cite{chowdhury2026ime}. However, predictive performance alone is insufficient for clinical decision support, where understanding the evidence underlying a model's prediction is important for interpretation and validation. Post-hoc explainability methods such as LIME~\cite{ribeiro2016lime}, SHAP~\cite{lundberg2017shap}, and Grad-CAM~\cite{selvaraju2020gradcam} address this need by identifying image regions or features associated with model predictions.

Most existing explanation methods produce a single explanation for a given prediction. For segment-based approaches such as LIME, the explanation additionally depends on how the image is partitioned into interpretable segments before their contributions are evaluated. Different segmentation strategies can produce substantially different segment boundaries and, consequently, different candidate evidence presented to the explanation method. Nevertheless, segmentation is commonly treated as a preprocessing choice, and its effect on the resulting explanation is rarely evaluated systematically.

Explanation stability presents an additional challenge. Perturbation- and sampling-based methods can produce different explanations across repeated evaluations of the same prediction~\cite{zhang2019lime_uncertainty,alvarezmelis2018robustness,chowdhury2026ime}. Such variability is particularly important in clinical applications, where explanations should provide reproducible evidence for model predictions. Existing approaches have primarily addressed this problem by improving the stability of an individual explanation. However, this still leaves open whether a single explanation adequately represents the different subsets of image segments that can support the same model prediction.

We formulate segment-based explanation as a multi-objective problem involving two competing requirements: preserving the classifier's prediction confidence and reducing the number of segments required to explain that prediction. Retaining additional segments can preserve more predictive information, whereas selecting fewer segments provides a more compact representation of the supporting evidence. Rather than imposing a predefined weighting between these objectives, multi-objective optimization can identify a set of non-dominated solutions representing different fidelity--compactness trade-offs.

Based on this formulation, we introduce \moxie{} (MultiObjective eXplanation Imaging Engine), an evolutionary framework for segment-based image explanations. Given a segmented image, \moxie{} represents each candidate explanation as a subset of segments and simultaneously optimizes prediction fidelity and explanation compactness. The resulting Pareto front provides multiple candidate explanations at different operating points, allowing the relationship between retained image evidence and preservation of model confidence to be examined explicitly.

A second objective of this work is to determine how image segmentation affects the explanations that can be found. Because MOXIE searches over segments, the segmentation defines its search space. We therefore evaluate MOXIE with four structurally different segmentation methods (Voronoi, SLIC, Watershed and Felzenszwalb) under identical objectives and experimental conditions. This controlled comparison shows how the way segments are formed changes both the explanations and how their quality is measured.

We evaluate MOXIE on two biomedical imaging datasets, BloodMNIST~\cite{yang2023medmnist} for blood-cell classification and HAM10000~\cite{tschandl2018ham10000} for skin-lesion classification, and compare it with LIME under the same budget of model evaluations. Results show that proposed MOXIE framework performed better than LIME and achieved a higher hypervolume on every test image. 
The comparison also exposes a limitation of LIME's single fixed-size explanation: its highlighted segments often fail to preserve the prediction when shown alone, whereas MOXIE’s alternatives reveal how much of the image the model needs and which context it relies on.
The main contributions of this work are:

\begin{itemize}
    \item We introduce MOXIE, a multi-objective evolutionary framework that jointly optimizes prediction fidelity and explanation compactness and returns a Pareto front of alternative explanations instead of a single explanation of predefined size.
    \item We show, under an equal evaluation budget, that MOXIE outperforms LIME on two biomedical datasets, and that its alternative explanations reveal how much image evidence a prediction requires, which a single explanation cannot show.
    \item We provide a controlled study of four segmentation methods, showing that segmentation shapes the resulting explanations and that counting segments can misrepresent compactness when segment sizes are unequal.
\end{itemize}

The remainder of this paper is organized as follows. Section~\ref{sec:related} reviews related work. Section~\ref{sec:method} presents the proposed framework. Section~\ref{sec:results} reports and discusses the experimental results, and Section~\ref{sec:conclusion} concludes the paper.

\section{Related Work}
\label{sec:related}

Post-hoc explanation methods provide different mechanisms for identifying evidence associated with model predictions~\cite{bhati2024survey}. LIME constructs a local surrogate model from perturbed samples to estimate the contributions of interpretable image segments~\cite{ribeiro2016lime}, while SHAP derives feature attributions based on Shapley values~\cite{lundberg2017shap}. Gradient-based approaches such as Grad-CAM use model activations and gradients to localize image regions associated with a target class~\cite{selvaraju2020gradcam}. Despite their methodological differences, these approaches typically summarize the explanation of an individual prediction using a single attribution map or set of important segments.

The reliability of such explanations has been widely investigated. Zhang~\cite{zhang2019lime_uncertainty} and Alvarez-Melis and Jaakkola~\cite{alvarezmelis2018robustness} showed that explanations can vary across repeated evaluations or small changes to the input. Krishna~\cite{krishna2022disagreement} further demonstrated that different explanation methods can disagree substantially regarding feature importance for the same prediction. Together, these findings raise concerns about treating an individual explanation as a unique representation of the evidence supporting a model's decision.

Previous work observed substantial variation across repeated LIME explanations for AD classification from MRI and introduced Influence Map based Explanation (\ime{})~\cite{chowdhury2026ime}. \ime{} aggregates information from repeated LIME evaluations to obtain a more stable representation and was evaluated using Confidence Impact Ratio (CIR) and Decision Impact Ratio (DIR). Although this approach reduces variability across repeated explanations, its objective remains the construction of a single consolidated explanation. It therefore does not characterize alternative segment subsets that may preserve the same prediction with different levels of compactness. This motivates a multi-objective formulation in which prediction fidelity and explanation size are optimized simultaneously and the resulting trade-offs are retained rather than collapsed into one explanation.


Multi-objective optimization has previously been used in explainable AI to represent competing requirements without collapsing them into a single weighted objective. Dandl~\cite{dandl2020multi} formulated counterfactual explanation generation as a multi-objective optimization problem and used an NSGA-II-based search~\cite{deb2002nsga2} to obtain diverse counterfactuals representing trade-offs among validity, proximity, sparsity, and plausibility. However, counterfactual explanations seek feature changes that alter the model output, whereas the present work seeks subsets of existing image evidence that preserve the original prediction. More closely related to image explanation, Wang~\cite{wang2022multiobjective} proposed a multi-objective genetic algorithm for evolving local model-agnostic explanations of deep image classifiers, demonstrating that evolutionary search can recover sets of non-dominated local explanations. Wang~\cite{wang2024multi} subsequently formulated feature-attribution explanation as a multi-objective learning problem, jointly considering faithfulness, sensitivity, and complexity and producing explanations with different trade-offs among these criteria. 

\begin{figure*}[t]
    \centering
    \includegraphics[width=14cm, height=6cm]{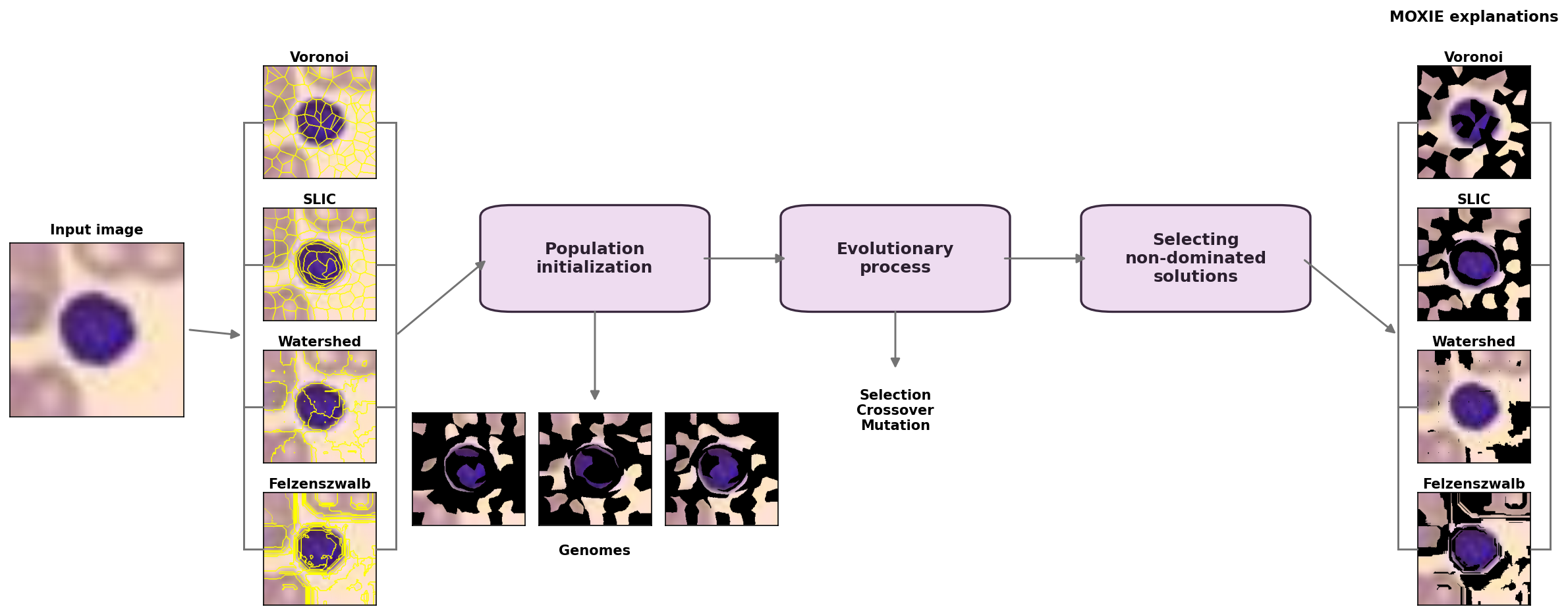}
    \caption{Overview of \moxie{}. The input image is divided into segments by one of four segmentation methods. Each candidate explanation is a genome with one gene per segment; segments whose gene exceeds $\tau = 0.7$ are kept and the rest are set to black.}
    \label{fig:moxie_framework}
\end{figure*}

Palatnik~\cite{de2022evolved} proposed EvEx for lymph-node metastasis classification, combining LIME with a multi-objective genetic algorithm to optimize image-segmentation parameters and aggregating explanations obtained from Pareto-optimal segmentations. More recently, Edgar~\cite{iglesias2025improving} used a multi-objective genetic algorithm to generate explanations for medical X-ray classification by searching for minimal changes in a learned image representation that alter the classifier output. Related evolutionary approaches have also used Pareto optimization to identify compact predictive representations in computational pathology~\cite{hashemian2026evops}. These studies demonstrate the utility of multi-objective search in biomedical imaging, but address different optimization problems. MOXIE instead searches directly over subsets of image segments that preserve the original predicted-class confidence while minimizing retained evidence, and retains the resulting non-dominated subsets as alternative explanations. It further evaluates how the underlying segmentation strategy changes the alternative explanation space.

Segment-based explanation methods require an image to be decomposed into interpretable units before the contribution of individual segments can be evaluated. The segmentation strategy therefore determines the candidate segments available to the explanation procedure. Because segmentation algorithms impose different assumptions about spatial proximity, appearance, gradients, and boundaries, their choice can affect both the composition and granularity of the resulting explanations.
SLIC~\cite{achanta2012slic} constructs compact superpixels based on spatial and appearance similarity. Felzenszwalb's graph-based method~\cite{felzenszwalb2004graphseg} generates adaptive segments by comparing graph-based measures of within- and between-segment variation. Watershed segmentation~\cite{vincent1991watershed} uses image gradients to identify segment boundaries, whereas centroidal Voronoi tessellation~\cite{du1999centroidal} partitions the image according to geometric proximity and therefore provides a useful content-independent baseline.

Despite these fundamentally different segment-generation mechanisms, segmentation is often fixed before explanation generation and is not evaluated as an experimental factor. This makes it difficult to determine whether observed explanation quality arises primarily from the attribution procedure or partly from the segments supplied to it. Motivated by calls for rigorous and task-grounded evaluation of interpretability methods~\cite{doshivelez2017rigorous}, we explicitly vary the segmentation strategy while holding the explanation objectives and evaluation procedure fixed. This enables direct assessment of segmentation as a source of variation in segment-based explanations.

\section{Proposed Methodology}
\label{sec:method}
\subsection{Multi-Objective eXplanation Imaging Engine Framework}

A segmentation-based explanation must balance two goals. It should keep the classifier's prediction, and it should show as little of the image as possible. These goals conflict: a smaller explanation usually preserves less of the prediction. Methods such as LIME resolve this conflict implicitly by returning a single explanation of fixed size. The user therefore never sees how the prediction changes as more or less of the image is shown. \moxie{} makes this trade-off explicit. It treats explanation as a bi-objective search and returns a set of explanations, each representing a different balance between faithfulness and compactness.

\begin{figure*}[t]
    \centering
    \subfloat[Original]{\includegraphics[width=3.2cm, height=3.2cm]{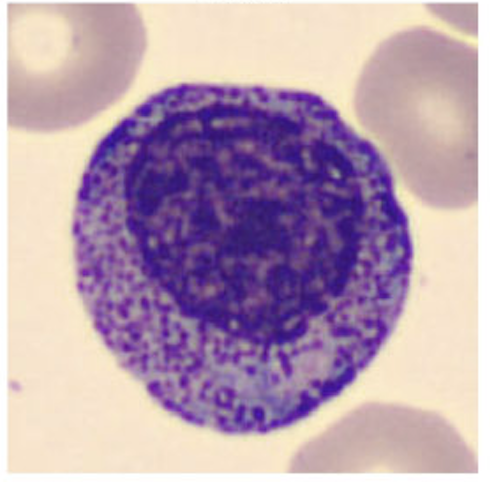}}
    \subfloat[Voronoi]{\includegraphics[width=3.2cm, height=3.2cm]{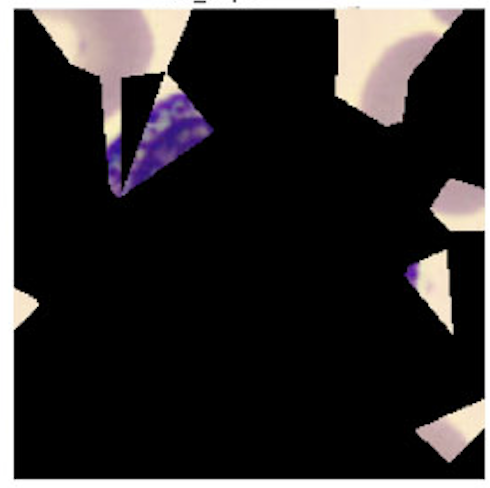}}
    \subfloat[SLIC]{\includegraphics[width=3.2cm, height=3.2cm]{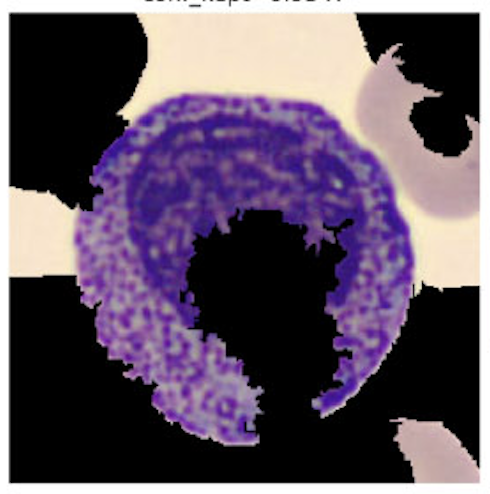}}
    \subfloat[Watershed]{\includegraphics[width=3.2cm, height=3.2cm]{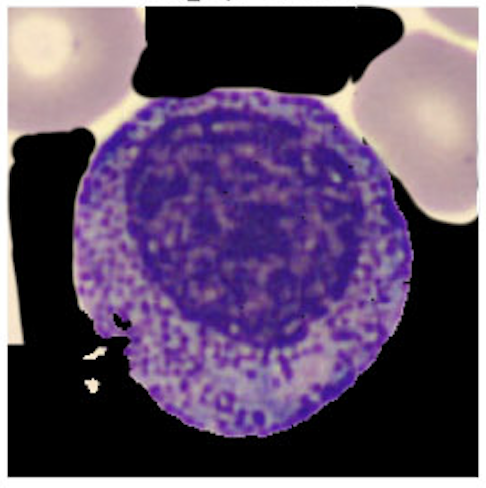}}
    \subfloat[Felzenszwalb]{\includegraphics[width=3.2cm, height=3.2cm]{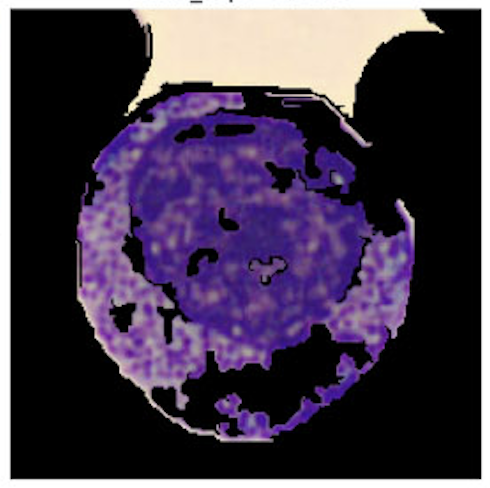}}
    \caption{A bloodMNIST image segmentation example with Voronoi, SLIC, Watershed and Felzenszwalb methods. Black segments are excluded from the explanation. Each segmentation method produces different segments, and therefore a different trade-off between faithfulness and compactness.}
    \label{fig:blood}
\end{figure*}

As shown in Fig.~\ref{fig:moxie_framework}, the input image $I$ is first divided into $K$ non-overlapping segments $S=\{S_1,\ldots,S_K\}$. A candidate explanation is encoded as a genome
\begin{equation}
\mathbf{x}=[x_1,\ldots,x_K], \qquad x_k\in[0,1],
\end{equation}
with one gene per segment. A threshold $\tau=0.7$ turns the genome into a binary mask $b_k=\mathbf{1}[x_k>\tau]$, so segment $S_k$ is kept when $b_k=1$. All other segments are set to black, giving the masked image $I_{\mathbf{x}}$. If no gene exceeds $\tau$, the segment with the largest gene is kept, so every genome yields a valid explanation. Since we use simulated binary crossover (SBX) and polynomial mutation, the genome is represented using real-valued variables and subsequently converted to a binary representation using a predefined threshold.

Let $\hat{c}$ be the class predicted for $I$, $p_{\mathrm{orig}}$ the classifier's confidence in $\hat{c}$ on $I$, and $p_{\mathrm{kept}}$ its confidence in $\hat{c}$ on $I_{\mathbf{x}}$. The first objective measures the deviation in predicted-class confidence caused by retaining only the selected segments and computed using eq.~\ref{eq:fidelity}.
\begin{equation}
f_1(\mathbf{x}) = |p_{\mathrm{kept}}-p_{\mathrm{orig}}|.
\label{eq:fidelity}
\end{equation}
The second objective measures the size of the explanation as the percentage of segments kept, as shown in eq.~\ref{eq:compactness}.
\begin{equation}
f_2(\mathbf{x}) = \frac{100}{K}\sum_{k=1}^{K} b_k.
\label{eq:compactness}
\end{equation}
Both objectives are minimized. We do not combine them into a weighted sum, because any choice of weights would fix the trade-off in advance. \moxie{} instead keeps the non-dominated solutions. The result is a Pareto front in which each solution is a different explanation of the same prediction, ranging from very compact to highly faithful. Algorithm~\ref{alg:moee} summarizes the procedure.

\begin{algorithm}[t]
\caption{MOXIE Framework}
\label{alg:moee}
\begin{algorithmic}
\Require Classifier $\mathcal{M}$, image $I$, segmentation method $\mathcal{S}$, number of segments $K$, threshold $\tau$, population size $P$, generations $G$
\Ensure Pareto front $\mathcal{PF}$ of explanations, masked images
\State $S \gets \mathcal{S}(I, K)$ \Comment{Divide $I$ into $K$ segments}
\State $\hat{c} \gets \arg\max \mathcal{M}(I)$;\; $p_{\mathrm{orig}} \gets \mathcal{M}(I)[\hat{c}]$
\State Initialize $X^{(0)} \in [0,1]^{P\times K}$ uniformly at random
\For{$g = 0$ to $G-1$}
    \State $B \gets \mathbf{1}[X^{(g)} > \tau]$
    \State $I_{\mathbf{x}} \gets \mathbf{I} \times B$
    \State $p_{\mathrm{kept}} \gets \mathcal{M}(I_{\mathbf{x}})[\hat{c}]$
    \State $f_1 \gets |p_{\mathrm{kept}} - p_{\mathrm{orig}}|$ \Comment{Fidelity loss}
    \State $f_2 \gets 100 \cdot \sum_k b_k / K$ \Comment{Explanation size}
    \State Rank  = NondominatedSort($\mathbf{I_{x}}$)
    \State $I_{\mathbf{offspring}} \gets$ Combination($\mathbf{I_{x}}$, Rank)
    \State $I_{\mathbf{x}} \gets$ Elitism($\mathbf{I_{x}}, \mathbf{I_{offspring}}$ )
\EndFor
\State $\mathcal{PF} \gets$ non-dominated solutions of the final population
\State \Return $\mathcal{PF}$ and the corresponding masked images
\end{algorithmic}
\end{algorithm}

The segments are the building blocks of every explanation, so the segmentation method defines what \moxie{} can search over. A method that produces segments of very different sizes changes both what an explanation can show and how its size is counted. We therefore evaluate four segmentation methods with different segment shapes: Voronoi~\cite{jones2005voronoi}, SLIC~\cite{achanta2012slic}, Watershed~\cite{vincent1991watershed}, and Felzenszwalb~\cite{felzenszwalb2004graphseg}. 
The objectives and optimization settings are held fixed across methods, allowing differences in the resulting fronts to be examined with respect to how the image is partitioned.

The framework uses NSGA-II~\cite{deb2002nsga2}, which ranks solutions by non-dominated sorting and keeps them spread along the front using crowding distance. Offspring are created with simulated binary crossover ($\eta=15$, probability 0.9) and polynomial mutation ($\eta=20$, probability 0.01). We measure the quality of a front with its hypervolume (HV), the area of objective space it dominates with respect to a reference point. A higher HV means the front is both closer to the ideal and better spread. Because NSGA-II is stochastic, each experiment is repeated over ten independent runs.

\begin{figure*}[t]
    \centering
    \subfloat[Lympocyte Sample]{\includegraphics[width=0.4\linewidth]{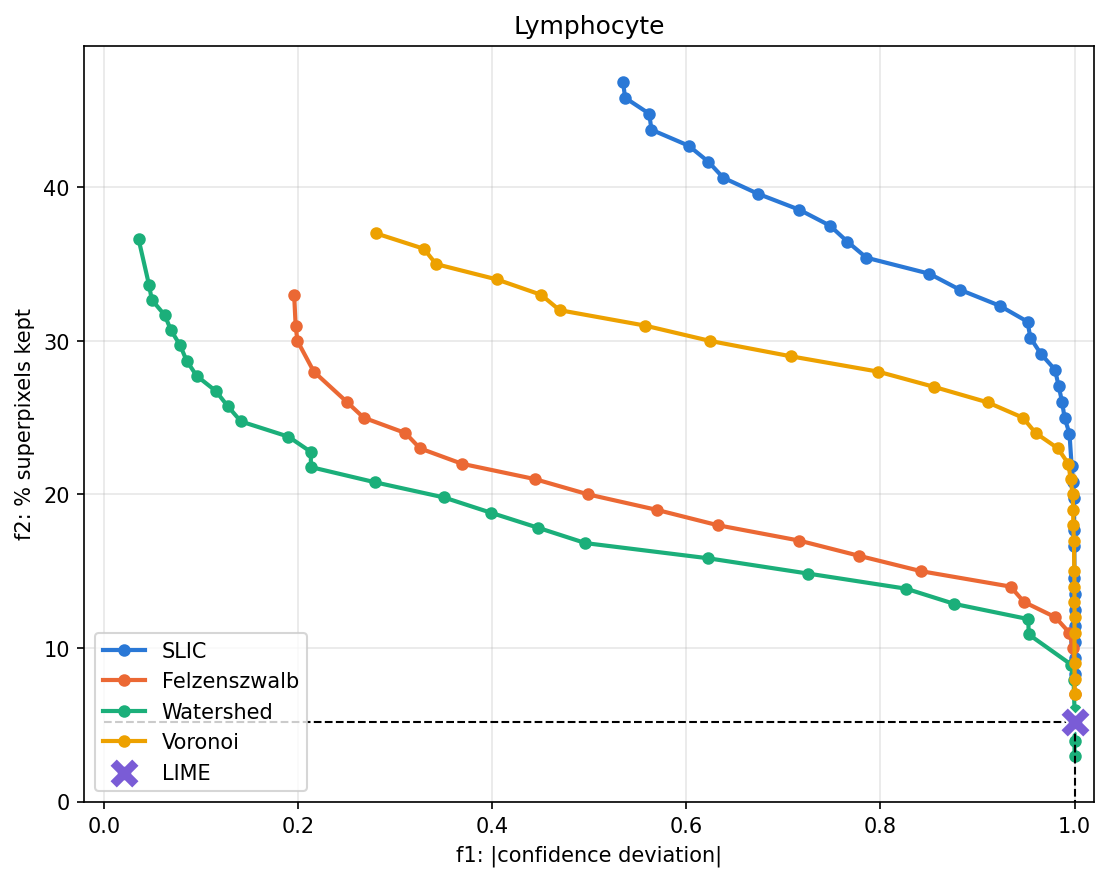}}
    \subfloat[Basal Cell Carcinoma Sample]{\includegraphics[width=0.4\linewidth]{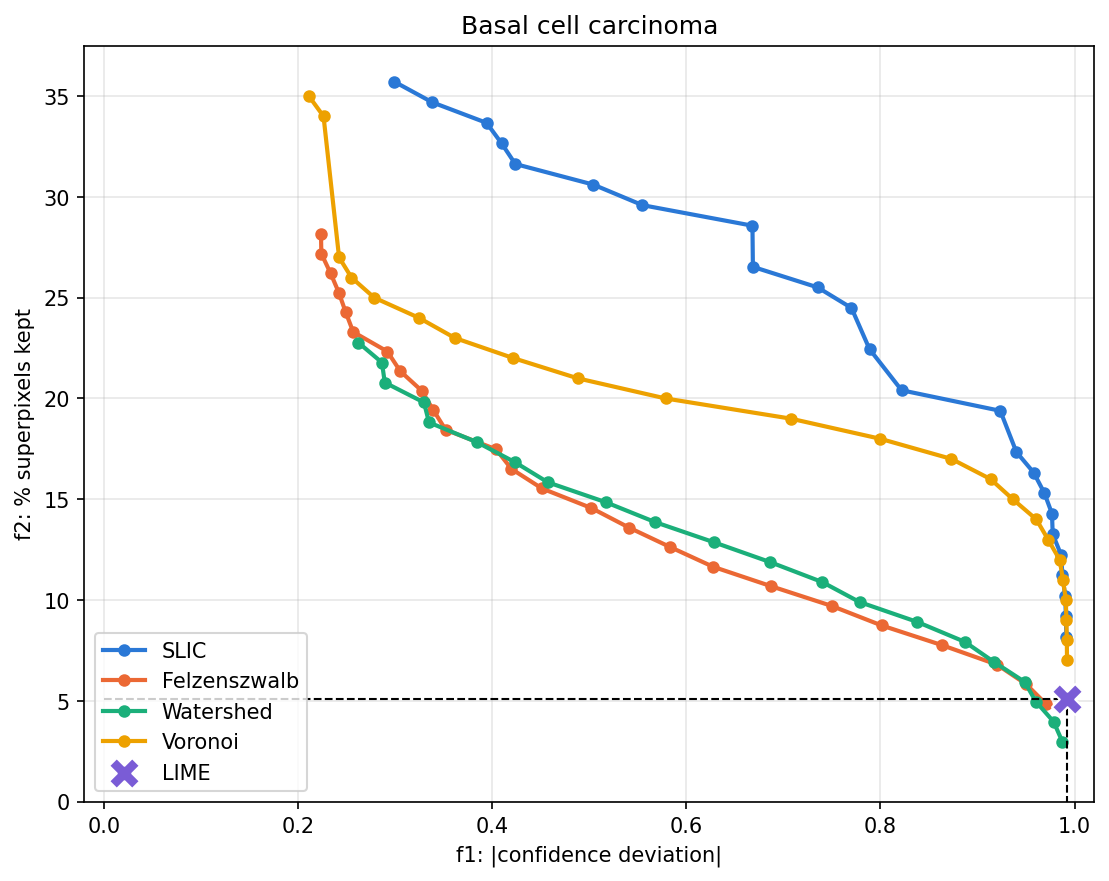}}
    \caption{Pareto fronts from the four segmentation methods and the LIME solution for a representative image from each dataset.}
    \label{fig:one_run}
\end{figure*}

\begin{figure*}
    \centering
    \includegraphics[width=0.8\linewidth]{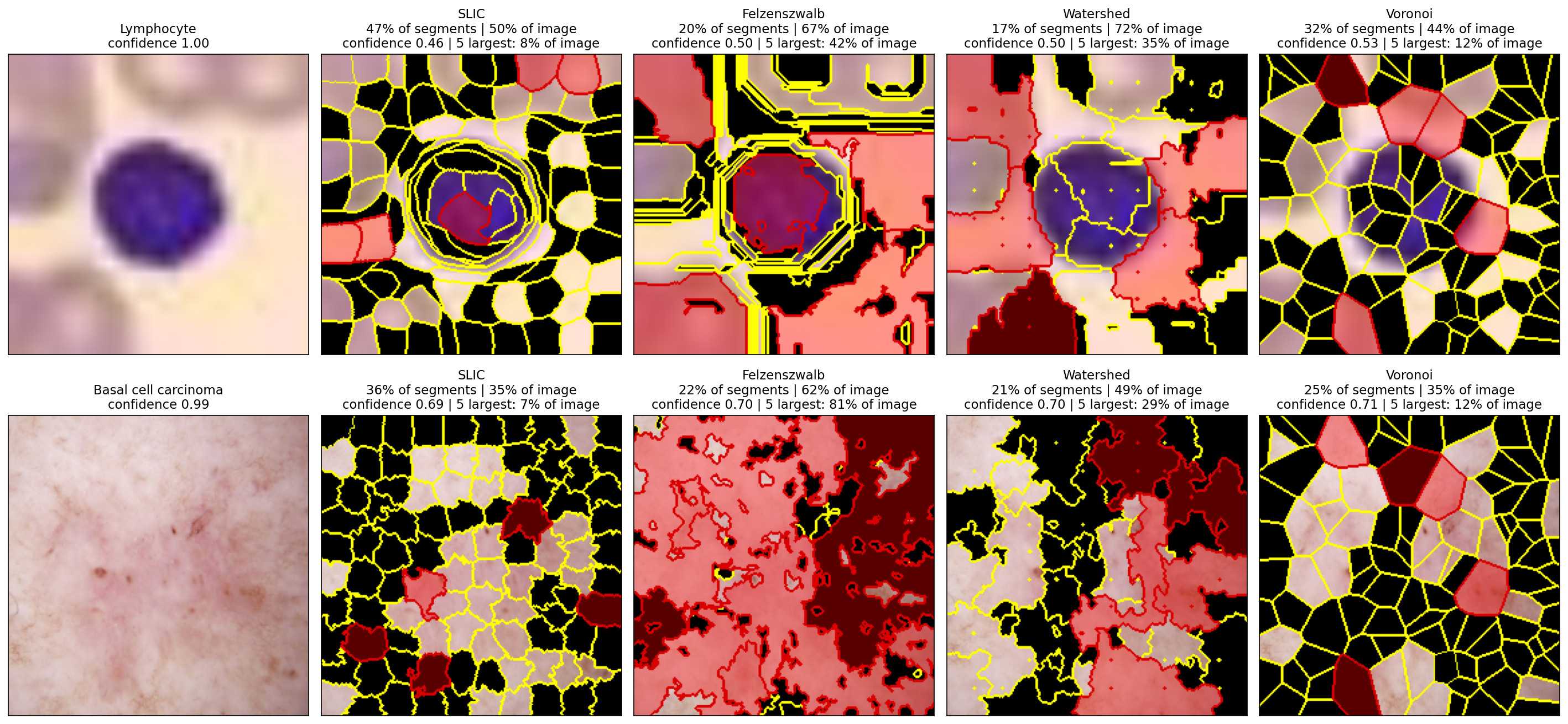}
    \caption{Comparison of explanations obtained using different segmentation methods. Red indicates the five largest retained segments by area. Watershed and Felzenszwalb retain fewer but larger segments, covering a greater portion of the image than SLIC and Voronoi in these examples.}
\label{fig:seg_matched}
\end{figure*}

\subsection{Datasets and Experimental Protocol}

Experiments are conducted on two biomedical imaging datasets: BloodMNIST~\cite{yang2023medmnist} and HAM10000~\cite{tschandl2018ham10000}. BloodMNSIT contains 17{,}092 blood-cell microscopy images in eight classes. We split the data into train/validation/test set as 11{,}959/1{,}712/3{,}421. HAM10000 contains 10{,}015 dermoscopic images of seven skin-lesion types, which we split 6{,}973/1{,}508/1{,}534 images at the lesion level, so that images of the same lesion never appear in more than one split. For every dataset, the classifier is an ImageNet-pretrained DenseNet121 fine-tuned for 15 epochs at $224\times224$ image size with Adam optimizer, learning rate of $10^{-4}$, batch size of 32, and step decay by 0.1 every 7 epochs.


For the experiments, 50 images were randomly sampled using a fixed seed from each test set to explain the classifier's predicted class. The sample size was selected to balance computational cost with evaluation across multiple images, segmentation methods, and independent runs.
Every image is partitioned into approximately 100 segments with each of four segmentation methods and then MOXIE uses NSGA-II with a population of 50 for 50 generations (2{,}500 model evaluations per run). A segment is active when its gene exceeds $\tau = 0.7$, and inactive segments are set to black. LIME~\cite{ribeiro2016lime} serves as the baseline and uses exactly the same segmentation and masking. It draws 2{,}500 perturbation samples per run, matching MOXIE's budget, and its explanation keeps the five segments with the highest positive weights. Because both methods are stochastic, each is run 10 times per image with different seeds, giving 50 images $\times$ 4 segmentation methods $\times$ 2 explanation methods (MOXIE and LIME) $\times$ 10 runs per dataset. Explanations are evaluated with two objectives: confidence deviation, $f_1(\mathbf{x}) = |p_{\mathrm{kept}}-p_{\mathrm{orig}}|$ for the predicted class $c$, and compactness, $f_2$, the percentage of segments kept. Both are minimized. The quality of each Pareto front is summarized by its hypervolume (HV), with $f_2$ rescaled to $[0,1]$ and reference point $(1,1)$; a LIME run's HV is that of its single solution. MOXIE and LIME are compared using the mean HV over the 10 runs per image. 

\begin{table}[t]
\centering
\caption{Classifier performance where DenseNet121 is fine-tuned for 15 epochs at $224\times224$ image size.}
\label{tab:classifiers}
\begin{tabular}{llcr}
\toprule
Dataset & Modality & Classes & Test acc. \\
\midrule
BloodMNIST & Blood-cell  & 8 & 97.87\% \\
HAM10000   & Dermoscopy  & 7 &  84.88\% \\
\bottomrule
\end{tabular}
\end{table}

\section{Results and Discussion}\label{sec:results}

This section evaluates MOXIE on BloodMNIST and HAM10000 and compares its explanations with LIME under a matched model-evaluation budget. We first report classifier performance, followed by quantitative evaluation of the fidelity–compactness trade-off and analysis of representative Pareto-optimal explanations across different segmentation strategies.

\begin{table*}[t]
\centering
\small
\caption{Quantitative comparison of MOXIE and LIME under a matched budget of 2,500 model evaluations per run. HV is reported as mean ± SD across images, with each image averaged over 10 independent runs. “Wins” denotes the number of images for which MOXIE achieves higher HV than LIME, and “LIME dom.” denotes the percentage of LIME solutions dominated by the MOXIE Pareto front. \(p_{\mathrm{Holm}}\) reports Holm-corrected statistical significance.}
\label{tab:moxie_vs_lime}
\begin{tabular}{llcccccc}
\toprule
Dataset & Segmentation & $n$ & MOXIE HV & LIME HV & Wins & LIME dom. & $p_{\text{Holm}}$ \\
\midrule
\multirow{4}{*}{BloodMNIST} & SLIC & 50 & 0.512 $\pm$ 0.298 & 0.107 $\pm$ 0.265 & 50/50 & 29\% & $1.4\times10^{-14}$ \\
 & Felzenszwalb & 50 & 0.669 $\pm$ 0.262 & 0.122 $\pm$ 0.269 & 50/50 & 36\% & $1.4\times10^{-14}$ \\
 & Watershed & 50 & 0.675 $\pm$ 0.259 & 0.121 $\pm$ 0.271 & 50/50 & 39\% & $1.4\times10^{-14}$ \\
 & Voronoi & 50 & 0.554 $\pm$ 0.281 & 0.108 $\pm$ 0.266 & 50/50 & 27\% & $1.4\times10^{-14}$ \\
\midrule
\multirow{4}{*}{HAM10000} & SLIC & 50 & 0.933 $\pm$ 0.139 & 0.750 $\pm$ 0.269 & 50/50 & 95\% & $1.4\times10^{-14}$ \\
 & Felzenszwalb & 50 & 0.933 $\pm$ 0.160 & 0.777 $\pm$ 0.234 & 50/50 & 97\% & $1.4\times10^{-14}$ \\
 & Watershed & 50 & 0.939 $\pm$ 0.146 & 0.787 $\pm$ 0.233 & 50/50 & 97\% & $1.4\times10^{-14}$ \\
 & Voronoi & 50 & 0.923 $\pm$ 0.160 & 0.746 $\pm$ 0.304 & 50/50 & 94\% & $1.4\times10^{-14}$ \\
\midrule
\multirow{4}{*}{\shortstack[l]{BloodMNIST\\(excl.\ erythroblast)}} & SLIC & 46 & 0.471 $\pm$ 0.274 & 0.035 $\pm$ 0.096 & 46/46 & 23\% & $2.3\times10^{-13}$ \\
 & Felzenszwalb & 46 & 0.642 $\pm$ 0.256 & 0.051 $\pm$ 0.115 & 46/46 & 30\% & $2.3\times10^{-13}$ \\
 & Watershed & 46 & 0.648 $\pm$ 0.253 & 0.049 $\pm$ 0.119 & 46/46 & 35\% & $2.3\times10^{-13}$ \\
 & Voronoi & 46 & 0.517 $\pm$ 0.261 & 0.035 $\pm$ 0.096 & 46/46 & 21\% & $2.3\times10^{-13}$ \\
\midrule
\multirow{4}{*}{\shortstack[l]{HAM10000\\(excl.\ \textit{nv})}} & SLIC & 12 & 0.772 $\pm$ 0.222 & 0.327 $\pm$ 0.189 & 12/12 & 81\% & $2.0\times10^{-3}$ \\
 & Felzenszwalb & 12 & 0.769 $\pm$ 0.273 & 0.490 $\pm$ 0.298 & 12/12 & 88\% & $2.0\times10^{-3}$ \\
 & Watershed & 12 & 0.789 $\pm$ 0.251 & 0.499 $\pm$ 0.298 & 12/12 & 90\% & $2.0\times10^{-3}$ \\
 & Voronoi & 12 & 0.741 $\pm$ 0.256 & 0.258 $\pm$ 0.182 & 12/12 & 80\% & $2.0\times10^{-3}$ \\
\bottomrule
\end{tabular}
\end{table*}

\begin{figure*}[t]
    \centering
    \includegraphics[width=0.8\linewidth]{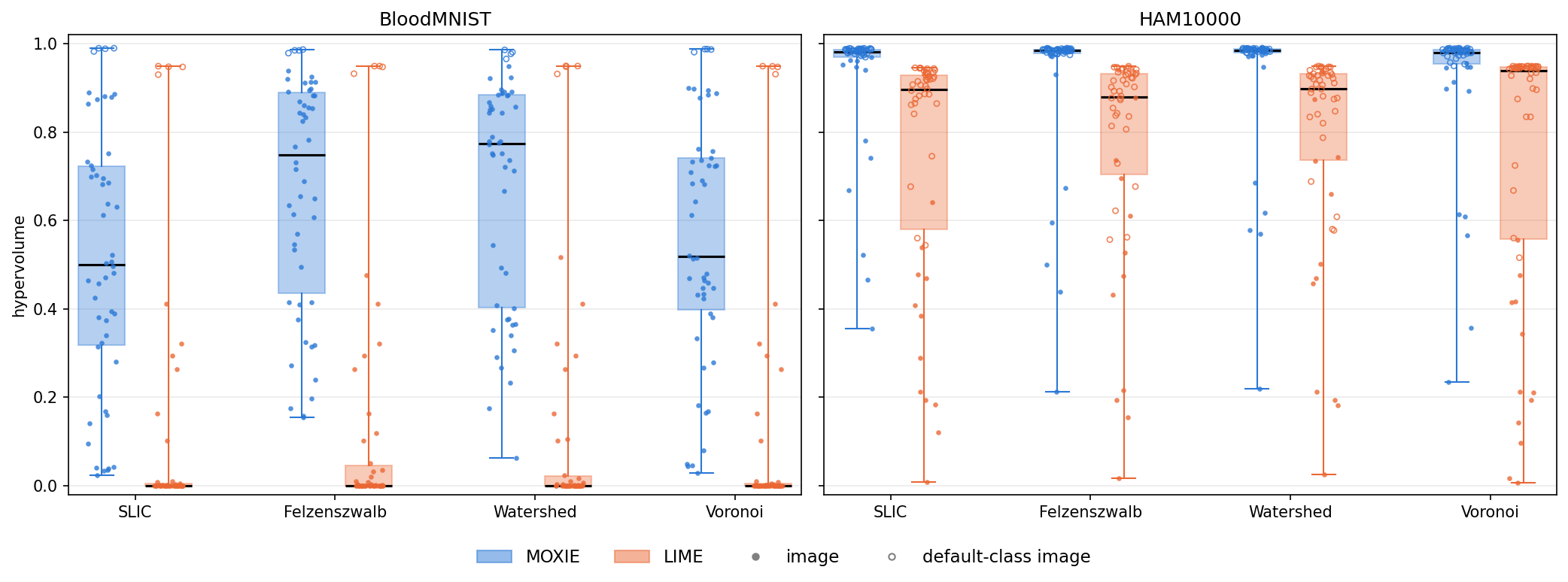}
    \caption{Distribution of hypervolume across 50 test images for MOXIE and LIME using SLIC, Felzenszwalb, Watershed, and Voronoi segmentation on BloodMNIST and HAM10000. Each value represents the mean HV over 10 independent runs for an image. Higher HV indicates a better fidelity–compactness trade-off.}
    \label{fig:box_hv}
\end{figure*}

Table~\ref{tab:classifiers} summarizes the performance of the fine-tuned DenseNet121 classifiers whose predictions are explained in the remainder of this section. On BloodMNIST, the model achieves a test accuracy of 97.87\%, whereas on HAM10000, where the training, validation, and test sets were split at the lesion level so that images of the same lesion never appear in more than one split, the model achieves 84.88\%. These results establish the predictive performance of the classifiers before their individual predictions are analyzed.

\begin{figure*}[t]
    \centering
    \includegraphics[width=0.8\linewidth]{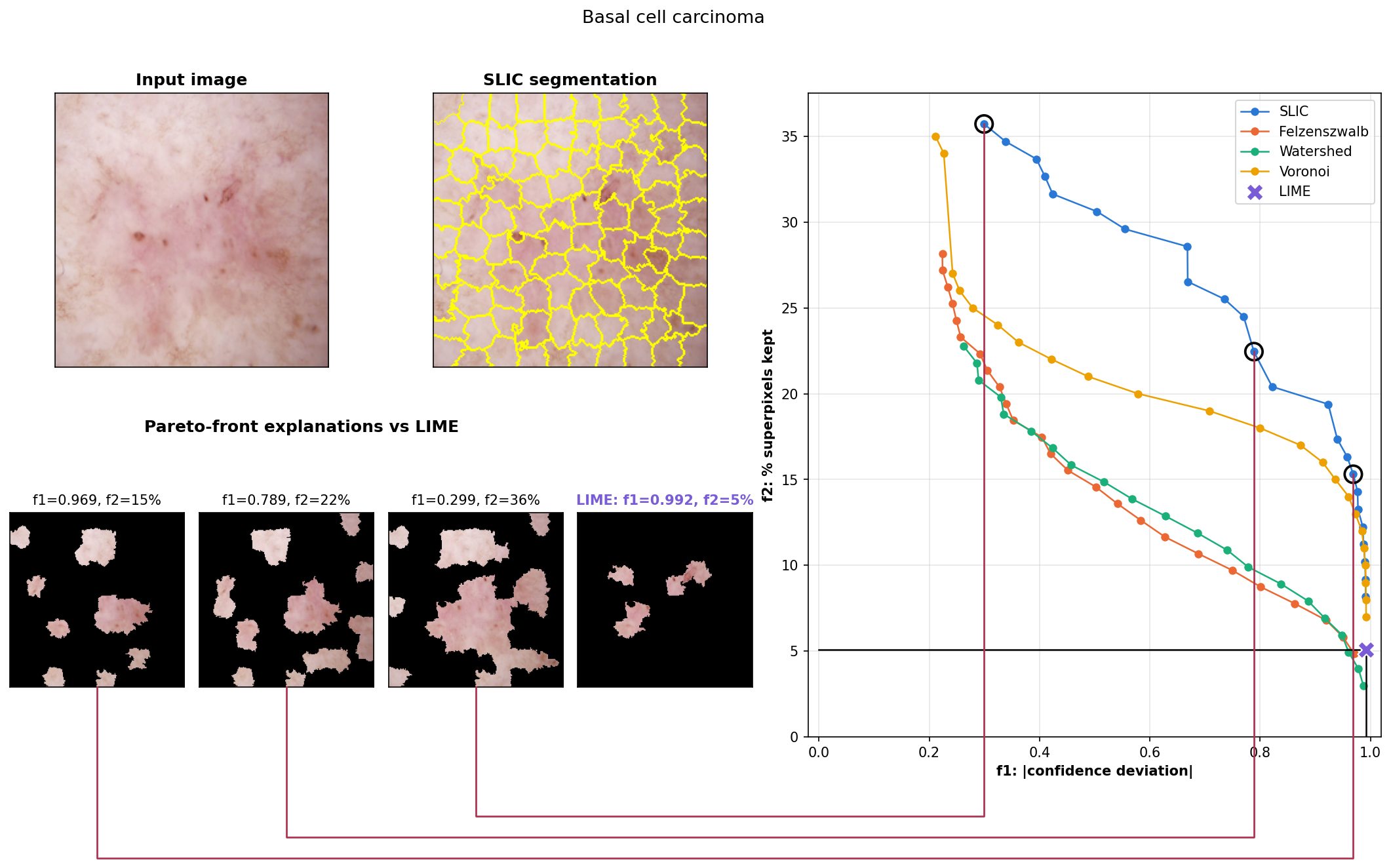}
    \caption{Pareto-optimal explanations found by \moxie{} for a HAM10000 basal cell carcinoma image (confidence 0.99), compared with LIME. Black regions indicate removed segments; percentages denote retained segments and \(f_1\) denotes confidence deviation.}
    \label{fig:f1}
\end{figure*}

\begin{figure*}[t]
    \centering
    \includegraphics[width=0.8\linewidth]{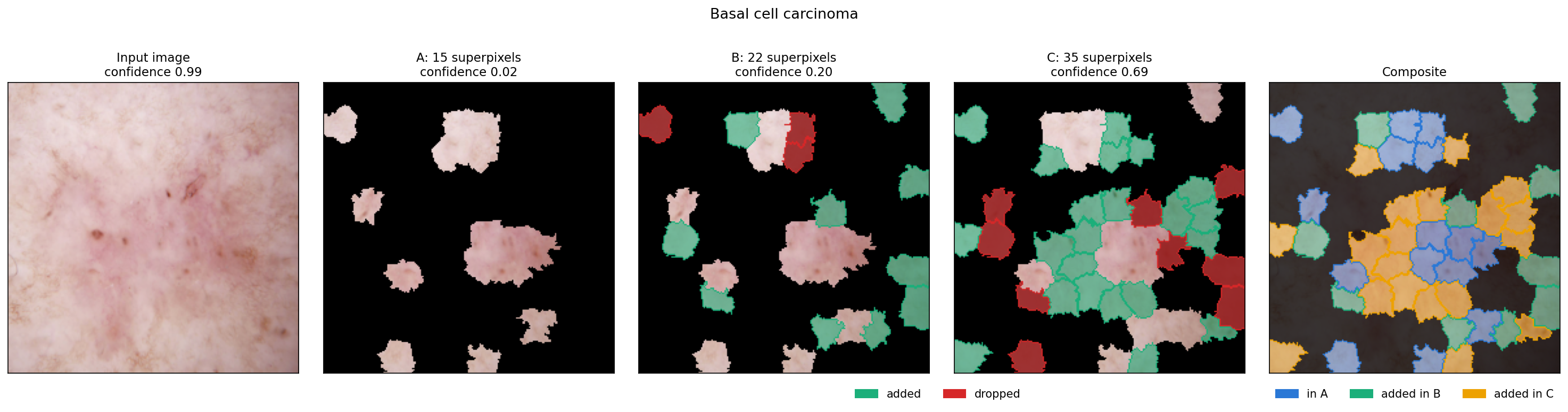}
    \caption{Comparison of three Pareto-optimal MOXIE explanations (A–C) from Fig.~\ref{fig:f1}, with predicted-class confidence shown for each. Green and red indicate regions added and removed, respectively, relative to the preceding explanation; the composite shows where regions first appear.}
    \label{fig:f3}
\end{figure*}

Figure~\ref{fig:one_run} illustrates a limitation of a single fixed-size explanation. For both representative images, a basal cell carcinoma from HAM10000 and a lymphocyte from BloodMNIST, LIME highlights about 5\% of the image segments. However, when the classifier is evaluated using only these regions, its predicted-class confidence falls from 0.99 to almost zero for the skin lesion and from 1.00 to zero for the blood cell. Thus, although the LIME explanations are compact, the selected regions alone do not preserve the original predicted-class confidence under the masking procedure. This loss of confidence is not apparent from the LIME explanation alone.

MOXIE exposes this trade-off by returning a set of alternative explanations rather than a single fixed-size explanation. Each front in Fig.~\ref{fig:one_run} contains explanations ranging from compact solutions with larger confidence deviation to larger solutions that more closely preserve the original predicted-class confidence. For the skin lesion, the model recovers a confidence of about 0.70–0.78 when roughly a quarter to a third of the segments are retained. For the blood cell, retaining about one-third of the regions with Watershed segmentation restores the confidence to 0.96. Thus, the Pareto front allows the fidelity–compactness trade-off to be examined directly rather than fixing the explanation size in advance.

The Pareto fronts also differ across segmentation methods for the same image, showing that the image partition changes the attainable fidelity–compactness trade-off. In these examples, Watershed and Felzenszwalb generally provide more favorable trade-offs when compactness is measured by the percentage of segments retained, allowing comparable confidence preservation with fewer segments than SLIC or Voronoi. This apparent advantage may arise from differences in segment geometry and size, since adaptive methods such as Watershed and Felzenszwalb can produce regions of unequal size, as shown in Fig.~\ref{fig:seg_matched}.  

Table~\ref{tab:moxie_vs_lime} compares MOXIE with LIME under a matched budget of 2,500 model evaluations per run, averaging each method over ten independent runs per image. MOXIE attains a higher hypervolume than LIME on every one of the 50 images for all four segmentation methods on both datasets. On BloodMNIST, the gap is large (0.512–0.675 vs. 0.107–0.122), with LIME's fixed selection of five superpixels producing a predicted-class confidence deviation greater than 0.9 in 76–80\% of images, indicating that these explanations are highly compact but frequently fail to preserve the original predicted-class confidence. On HAM10000, the overall gap is smaller (0.923–0.939 vs. 0.746–0.787). This is largely because 38 of the 50 images are predicted as nv and retain near-full confidence even when only about 1\% of the image is retained, resulting in high hypervolume for both methods. To examine whether these cases were driving the overall comparison, we repeated the analysis after excluding the nv predictions and the four BloodMNIST images predicted as erythroblast that exhibited similar behavior. The HAM10000 difference then increased substantially (0.741–0.789 vs. 0.258–0.499), with MOXIE obtaining higher HV for all 12 remaining images (\(p=0.002\)); the direction of the BloodMNIST result was likewise unchanged across all 46 remaining images.

Figure~\ref{fig:box_hv} shows the distribution of hypervolume values across these 50 test images for each segmentation method. On BloodMNIST, MOXIE generally achieves substantially higher HV than LIME, although the spread across images is relatively large, indicating that the attainable fidelity–compactness trade-off varies considerably between individual predictions. The separation between the two methods is observed across all four segmentation strategies, with Watershed and Felzenszwalb showing the highest average MOXIE HV. On HAM10000, both methods achieve higher HV overall and the distributions are more concentrated near high values. This behavior is largely associated with the predominance of images predicted as nv, for which high predicted-class confidence can be retained with very small explanations.

Figure~\ref{fig:f1} illustrates the difference between the two methods on a representative HAM10000 image correctly classified as basal cell carcinoma (confidence 0.99). LIME returns a single explanation of five superpixels, but the model's confidence on this explanation alone falls almost to zero ($f_1 = 0.992$): the explanation is compact but does not preserve the prediction. MOXIE instead returns a front of explanations spanning a range of trade-off between compactness and faithfulness. The three solutions shown, taken from the median-hypervolume SLIC run, keep 15\%, 22\% and 36\% of the superpixels and reduce $f_1$ to 0.969, 0.789 and 0.299, respectively. Fronts obtained with the other three segmentation methods show the same pattern: every front starts near LIME's solution and extends towards substantially more faithful explanations.

Figure~\ref{fig:f3} examines three solutions, A, B, and C, from the Pareto front in more detail. Solution A retains 15\% of the segments, primarily covering the central lesion, but the predicted-class confidence is only 0.02. Adding additional regions increases the confidence to 0.20 in B and 0.69 in C. Importantly, these solutions are not simply nested: B and C add new regions while also removing some regions retained by earlier solutions, demonstrating that the Pareto front contains genuinely different combinations of image regions rather than progressively larger versions of a single explanation. The higher-confidence solutions also include surrounding skin and image-border regions, indicating that the classifier's confidence is sensitive to regions beyond the central lesion. This behavior would not be apparent from a single fixed-size explanation such as LIME's.

\section{Conclusion}\label{sec:conclusion}

This work investigated whether a single segment-based explanation is sufficient to characterize the image evidence associated with a classifier's prediction. MOXIE formulates explanation as a multi-objective search over image-region subsets, producing alternatives that trade predicted-class confidence preservation against compactness. The results demonstrate that explanation multiplicity is not simply a matter of explanation size; different region combinations can characterize different fidelity--compactness trade-offs for the same prediction.
Experiments on BloodMNIST and HAM10000 further show that segmentation directly influences the explanations that can be discovered. With the classifier, objectives, and optimization procedure fixed, different segmentation strategies produce different fidelity--compactness fronts. Under a matched evaluation budget, MOXIE achieves higher hypervolume than LIME across all evaluated segmentation methods and images. These findings suggest that an image prediction may be better characterized by alternative region combinations and their trade-offs rather than by a single explanation.
Future work will investigate whether these alternatives correspond to semantically or clinically distinct evidence and evaluate them with domain experts.

\bibliographystyle{IEEEtran}
\bibliography{main}

\end{document}